\documentclass[conference]{IEEEtran}
\usepackage{cite}
\usepackage{amsmath,amssymb}
\usepackage{graphicx}
\usepackage{booktabs}
\usepackage{microtype}
\usepackage{url}
\newcommand{\aAblAftrMaeP}{0.376}

\newcommand{\aAblAftrSrP}{0.308}
\newcommand{\aAblBothMaeP}{0.474}

\newcommand{\aAblBothSrP}{0.781}

\newcommand{\aAblRegMaeP}{0.211}

\newcommand{\aAblRegSrP}{0.211}

\newcommand{\aAftDmZeroP}{0.710}
\newcommand{\aAftMae}{3.15}
\newcommand{\aAftSr}{0.28}
\newcommand{\aDlinDmZeroP}{0.009}
\newcommand{\aGruDmZeroP}{0.013}
\newcommand{\aLstmDmZeroP}{0.016}
\newcommand{\aMeanDa}{57.2}
\newcommand{\aMeanMae}{3.08}
\newcommand{\aMeanSr}{0.99}
\newcommand{\aNTest}{409}

\newcommand{\aTestEnd}{2025-12-22}
\newcommand{\aTestStart}{2024-05-07}
\newcommand{\aTrfMae}{3.41}

\newcommand{\bufBSr}{0.37}
\newcommand{\bufBSrt}{0.14}
\newcommand{\bufBTurn}{1.29}
\newcommand{\bufQSr}{0.29}
\newcommand{\bufQSrt}{$-$0.05}
\newcommand{\bufQTurn}{2.47}

\newcommand{\csNvdaAbsZLevel}{14.24}
\newcommand{\csNvdaAbsZStat}{0.75}
\newcommand{\gAftEntropy}{0.9982}
\newcommand{\gAftFinQk}{0.020}
\newcommand{\gAftMaxGate}{0.105}
\newcommand{\gAftMinGate}{0.078}

\newcommand{\gAftWeightSum}{0.091}
\newcommand{\gAftrEntropy}{0.9628}
\newcommand{\gAftrFinQk}{0.249}

\newcommand{\gUniform}{0.0909}
\newcommand{\hFIc}{0.0172}
\newcommand{\hFIcYa}{0.018}
\newcommand{\hFIcYb}{0.009}
\newcommand{\hFIcYc}{0.025}
\newcommand{\hFP}{0.038}
\newcommand{\hFT}{1.77}
\newcommand{\hTIc}{0.0003}
\newcommand{\hTIcYa}{$-$0.022}
\newcommand{\hTIcYb}{$-$0.010}
\newcommand{\hTIcYc}{0.036}
\newcommand{\hTP}{0.49}
\newcommand{\hTT}{0.02}

\newcommand{\hfXgbIc}{0.0213}

\newcommand{\hfXgbT}{2.29}

\newcommand{\htAftIc}{$-$0.0189}

\newcommand{\htMomIc}{0.0177}

\newcommand{\htRidgeIc}{0.0179}

\newcommand{\htRidgeSr}{0.46}

\newcommand{\htTrfIc}{$-$0.0178}

\newcommand{\htXgbIc}{$-$0.0213}

\newcommand{\sfMinP}{0.067}

\newcommand{\sfRidgeIc}{0.0123}

\newcommand{\sfRidgeT}{2.02}
\newcommand{\sfVixP}{0.52}

\newcommand{\sfXgbIc}{0.0155}

\newcommand{\sfXgbT}{2.00}

\newcommand{\spMax}{484}
\newcommand{\spMed}{369}
\newcommand{\spMin}{268}
\newcommand{\spN}{500}
\newcommand{\spValid}{1{,}390{,}254}
\newcommand{\stAftIc}{0.0332}
\newcommand{\stAftMdd}{$-$46}
\newcommand{\stAftSr}{0.36}
\newcommand{\stAftT}{1.81}
\newcommand{\stAftVsTrfD}{+0.0013}
\newcommand{\stAftVsTrfP}{0.869}
\newcommand{\stAftrIc}{0.0321}

\newcommand{\stRidgeIc}{0.0296}

\newcommand{\stRidgeT}{2.75}
\newcommand{\stTrfIc}{0.0319}

\newcommand{\stXgbIc}{0.0345}
\newcommand{\stXgbMdd}{$-$18}
\newcommand{\stXgbSr}{0.55}
\newcommand{\stXgbT}{2.54}

\newcommand{\sysfAdevSrf}{0.39}

\newcommand{\sysfAholdMdd}{$-$11}
\newcommand{\sysfAholdSrf}{0.34}

\newcommand{\sysfAholdSrz}{0.64}

\newcommand{\sysfFullholdExp}{65}

\newcommand{\sysfFullholdSrf}{0.62}
\newcommand{\sysfFullholdSrt}{0.32}

\newcommand{\systAdevSrf}{0.59}

\newcommand{\systAholdSrf}{0.08}
\newcommand{\systAholdSrt}{$-$0.02}

\newcommand{\tfXgbIc}{0.0059}

\newcommand{\tfXgbT}{0.78}

\newcommand{\releasestatement}{Code and the pre-registration are available at \url{https://github.com/Dishan18/FinancialTransformer} (archived at \url{https://doi.org/10.5281/zenodo.21109641}); all out-of-sample predictions are attached to its release v2.0.0.}

\begin{document}
\title{Does Regime-Gated Attention Help Stock Return Prediction?\\A Leakage-Controlled, Pre-Registered Evaluation}
\author{\IEEEauthorblockN{Dishan Sarkar}
\IEEEauthorblockA{Department of Artificial Intelligence and Machine Learning\\
Birla Institute of Technology, Mesra, Ranchi, India\\
btech10614.23@bitmesra.ac.in \quad ORCID: 0009-0002-8672-6680}}
\maketitle


\begin{abstract}
Regime-aware attention is an appealing inductive bias for stock-return prediction, but claims about it are rarely tested under leakage-controlled, out-of-sample conditions. We re-examine the Adaptive Financial Transformer (AFT), which biases self-attention with feature-group similarities weighted by gates that a market-regime encoder produces. First, on a single stock under a corrected protocol (aligned and purged windows, identical test dates for all models, cost-aware non-overlapping backtests, five seeds, Diebold--Mariano tests), no learned model beats the historical-mean forecast. The AFT's regime and context components are numerically inert: the learned gates have \gAftEntropy{} of maximal entropy, the context bias has \gAftFinQk{} of the scale of the attention logits, and we show that the product of two simplex-normalised weightings attenuates it by a factor $1/G$. Removing these components changes nothing. Second, we build a point-in-time S\&P~500 panel (\spN{} stocks, \spValid{} stock-days) with 63 stationary features whose causality is verified exactly, and evaluate a model zoo and a layered meta-model with yearly walk-forward refits, a development period (2016--2022) and a sealed, pre-registered holdout (2023--2025). Cross-sectional breadth turns weak signals into detectable ones; the AFT is statistically indistinguishable from a vanilla Transformer and less stable than gradient boosting or ridge regression; six learned meta-learners fail to beat an equal-weight rank combination; and a hysteresis rule roughly halves turnover. On the holdout, the pre-registered primary system (21-day horizon) fails (rank IC \hTIc, one-sided $p=\hTP$), whereas the same system at a 5-day horizon, a pre-registered secondary result, attains a rank IC of \hFIc{} (one-sided $p=\hFP$) and a net long-short Sharpe ratio of \sysfAholdSrf{} at 5\,bp per side. \releasestatement{}
\end{abstract}


\section{Introduction}
Daily stock returns have a very low signal-to-noise ratio, and their dynamics change with market conditions. Both facts make regime-aware deep architectures attractive. They also make empirical claims fragile: small evaluation choices (window alignment, label overlap at split boundaries, which dates each model is tested on, transaction costs, the number of seeds, and how often one looks at the test set) can move results by more than the effect being measured~\cite{deprado2018,bailey2014,harvey2016}.

This paper studies one such architecture, the Adaptive Financial Transformer (AFT). The AFT conditions self-attention on a learned market regime: a regime encoder summarises the input window, a gate network turns the regime into weights over 11 semantic feature groups, and the weighted group similarities are added to the attention logits. We ask three questions. \emph{Does the regime-gated attention help?} \emph{If not, why not?} \emph{And where, if anywhere, does a regime- or model-aware design add value to a stock-return forecaster?}

\textbf{Contributions.}
(i)~A mechanistic diagnosis. The AFT's gate acts on a product of two probability vectors, which attenuates its effect by $1/G$. In trained models the gates are almost exactly uniform and the context bias is negligible. Ablations confirm that the AFT's specific components do not affect its forecasts, and repairing the attenuation does not change performance (Sec.~\ref{sec:aft}).
(ii)~A leakage-controlled evaluation design for stock-return models: stationary, rank-normalised features with an exact look-ahead test; a point-in-time S\&P~500 universe; nested purged walk-forward evaluation; a development/holdout split with a written pre-registration; and tests that account for overlapping targets (Sec.~\ref{sec:protocol}).
(iii)~Evidence on where value comes from: breadth rather than architecture; equal weighting rather than learned meta-learning; cost-aware execution (Sec.~\ref{sec:results}).
(iv)~An honest out-of-sample record. The system chosen on the development period failed on the sealed holdout, while its 5-day counterpart survived (Sec.~\ref{sec:holdout}).

\section{Related Work}
\label{sec:related}
\textbf{Attention models for time series and stocks.} The Temporal Fusion Transformer learns variable-selection weights~\cite{lim2021}; Informer, PatchTST and iTransformer change the attention pattern~\cite{zhou2021,nie2023,liu2024}; and a decomposition-linear model often matches them~\cite{zeng2023}. Relative-position representations in self-attention go back to Shaw et al.~\cite{shaw2018}. For stocks, MASTER uses a market-status vector to rescale features and attends across stocks~\cite{li2024master}, which is closely related to the AFT's regime gating; earlier work used recurrent and attention models on daily data~\cite{fischer2018,ding2020}. In large-scale asset pricing, tree ensembles and neural networks add predictive power, but out-of-sample $R^2$ is small~\cite{gu2020}.

\textbf{Forecast combination and meta-learning.} Equal-weight combinations are hard to beat (the ``forecast combination puzzle''), including for equity returns~\cite{rapach2010}. Learned, feature-based weights work for collections of time series~\cite{montero2020}. We test both ideas, with the market regime and each model's realised skill as meta-features.

\textbf{Anomalies and regimes.} Short-term reversal~\cite{jegadeesh1990,lehmann1990} earns more when liquidity is scarce and volatility high~\cite{nagel2012}. Momentum~\cite{jegadeesh1993}, idiosyncratic volatility~\cite{ang2006} and lottery demand~\cite{bali2011} serve as factor baselines. By the fundamental law of active management, risk-adjusted performance scales with breadth~\cite{grinold2000}.

\textbf{Evaluation.} Purging, embargoes and backtest overfitting are discussed in~\cite{deprado2018,bailey2014}, data snooping in~\cite{white2000,harvey2016}. We compare forecasts with the Diebold--Mariano test~\cite{diebold1995}, with the Harvey--Leybourne--Newbold correction~\cite{harvey1997}, and use Newey--West standard errors~\cite{newey1987} throughout.

\section{The Adaptive Financial Transformer and Why Its Gate Is Inert}
\label{sec:aft}
\textbf{Architecture.} The input is a window $X_t\in\mathbb{R}^{L\times D}$ ending at the forecast origin~$t$, whose $D$ features are partitioned into $G$ groups $\mathcal{I}_g$. A pre-norm Transformer encoder with a classification (CLS) token produces the forecast from the CLS state. The \emph{regime encoder} embeds each group at every time step with a two-layer MLP, pools over time with learned attention and fuses the group summaries into a regime vector $R$. The \emph{gate network} maps $R$ to gates $w=\operatorname{softmax}(\mathrm{MLP}(R))\in\Delta^{G-1}$. In every layer, each group is projected to $\Phi_g\in\mathbb{R}^{(L+1)\times d_k}$ (a zero row stands for the CLS position), its rows are $\ell_2$-normalised, and $M_g=\hat\Phi_g\hat\Phi_g^\top\in[-1,1]^{(L+1)\times(L+1)}$. The financial-context bias and the attention of head $\eta$ are
\begin{align}
B_{\mathrm{fin}}&=\sigma(\phi)\textstyle\sum_{g=1}^{G}\pi_g\,w_g\,M_g,\qquad \pi=\operatorname{softmax}(\theta),\label{eq:bfin}\\
A_\eta&=\operatorname{softmax}\big(Q_\eta K_\eta^\top/\sqrt{d_k}+B_{\mathrm{fin}}+\sigma(\psi)\,T_\eta\big),\label{eq:attn}
\end{align}
where $\pi$ are learnable static group weights (initialised uniform), $\sigma$ is the logistic function and $T_\eta$ is a learnable relative-position bias.

\textbf{Attenuation.} Because $\pi$ and $w$ both lie on the probability simplex, $\sum_g\pi_g w_g$ is a convex combination of the gates, so $\sum_g\pi_gw_g\le\max_g w_g$, and it equals $1/G$ whenever either vector is uniform. Since $|(M_g)_{ij}|\le1$,
\begin{equation}
|(B_{\mathrm{fin}})_{ij}|\le\sigma(\phi)\textstyle\sum_g\pi_gw_g,\qquad
\Big|\frac{\partial (B_{\mathrm{fin}})_{ij}}{\partial w_g}\Big|\le\sigma(\phi)\,\pi_g .
\label{eq:atten}
\end{equation}
At initialisation ($\pi_g=1/G$) both the bias and its sensitivity to the gates are therefore $O(1/G)$. The gradient that would teach the regime encoder to discriminate between groups is attenuated by the same factor. With $G=11$ and $\sigma(0.5)=0.62$, the bias is at most $0.057$ times the mean group similarity. A one-line repair, \emph{AFT-R}, renormalises the effective weights, $\tilde w_g=\pi_gw_g/\sum_k\pi_kw_k$, which removes the $1/G$ factor without adding parameters.

\textbf{Measured on trained models} (AAPL, five seeds, test windows): the mean gate entropy is \gAftEntropy{} of its maximum, the mean smallest and largest gates are \gAftMinGate{} and \gAftMaxGate{} (uniform: \gUniform), $\sum_g\pi_gw_g=\gAftWeightSum$, and the standard deviation of $B_{\mathrm{fin}}$ is \gAftFinQk{} times that of the query--key logits. AFT-R raises the bias to \gAftrFinQk{} of the logit scale and makes the gates less uniform (entropy \gAftrEntropy). Table~\ref{tab:aapl} shows that neither the original nor the repaired gate matters for the forecasts.

\section{Data and Protocol}
\label{sec:protocol}
\textbf{Single-stock setting.} Daily AAPL data (2015--2025, Yahoo Finance, split- and dividend-adjusted), 95 technical features and the forward 5-day return. Windows end at the forecast origin; the last $h$ training and validation origins are purged; windows may draw features from an earlier split, so all models predict the same \aNTest{} test origins (\aTestStart{} to \aTestEnd). Every neural model uses the same training loop (AdamW, early stopping on validation MSE, five seeds, deterministic GPU kernels where PyTorch provides them). We test forecasts with the Diebold--Mariano test on the seed-averaged forecast. Trading is evaluated with a sign strategy on non-overlapping 5-day holding periods, averaged over the five possible rebalancing phases, with costs of 5\,bp per side on traded notional.

\textbf{Point-in-time panel.} We use the \spN{} current S\&P~500 constituents with usable Yahoo Finance histories (2010--2025). A stock enters the universe only on its index-addition date, which gives \spMin--\spMax{} members per date from 2011 onward (median \spMed). Prices before inclusion are used only to compute that stock's own features. Six series whose adjusted prices jump on a corporate-action date (e.g.\ the Fortive spin-off of DHR) start at that date. Stocks deleted from the index before 2026 are missing, a survivorship bias we state as a limitation.

\textbf{Features v2.} Of the original 95 features, 48 are price-denominated or cumulative quantities (prices, moving averages, dollar ranges, OBV, VWAP). On NVDA their test-period values have a mean absolute $z$-score of \csNvdaAbsZLevel{} under training-period scaling, against \csNvdaAbsZStat{} for the stationary features, which illustrates why such features generalise poorly. We therefore re-engineered 63 stationary, scale-free features: multi-horizon log returns and 12-1 momentum; volatility-normalised returns; realised, Parkinson, Garman--Klass and downside volatility; intraday range, close location and gap; log price-to-moving-average ratios, MACD/price and RSI; distance to the 21- and 252-day high and low; volume $z$-scores, relative and dollar volume and Amihud illiquidity; rolling skewness, kurtosis and extreme returns; beta, idiosyncratic volatility and correlation with the S\&P~500 ETF; returns relative to the leave-one-out sector mean; 11 market-state features (index returns, volatility and trend, VIX level and change, a variance-risk-premium proxy, cross-sectional dispersion and breadth); and 4 calendar terms. Stock features are rank-normalised across the universe on each date; market features are $z$-scored with training statistics. \emph{Look-ahead test:} we recompute all features with every input truncated at random dates; they equal the full-sample features exactly.

\textbf{Targets and evaluation.} The target is the forward $h$-day return in excess of the cross-sectional mean ($h\in\{5,21\}$). For each model and date we compute the rank IC (Spearman correlation between forecasts and realised returns across stocks) and report its mean with a Newey--West $t$-statistic (lag $h-1$). Long-short portfolios buy the top and sell the bottom quintile with equal weights, hold them for $h$ days, rebalance every $h$ days and average all statistics over the $h$ rebalancing phases. Costs are charged per side on traded notional.

\textbf{Walk-forward design.} Models are refitted every year on an expanding window (validation: the last 252 dates; $h$-day purges), producing out-of-sample forecasts for 2016--2025. \emph{Development period:} 2016--2022, on which every design decision was made. \emph{Holdout:} 2023--2025, sealed until a written pre-registration fixed the systems and hypotheses; it was then evaluated once.

\textbf{Model zoo.} Ridge regression (penalty chosen by validation IC), XGBoost~\cite{chen2016} (500 trees, depth 6, early stopping) and an MLP (32-16-8) act on the current feature vector. A vanilla Transformer, the AFT and AFT-R act on 20-day windows ($d=64$, 4 heads, 2 layers) and are trained on every fifth date, because consecutive targets overlap. All neural models use three seeds, MSE on winsorised targets and early stopping on validation rank IC. Classic factors (5- and 21-day reversal, 12-1 momentum, low volatility, MAX) serve as zero-parameter baselines.

\textbf{Meta-model.} The meta level acts only on out-of-sample forecasts, converted to daily cross-sectional ranks. Meta-learners are refitted yearly on earlier out-of-sample dates, and every input at date $t$ (ranks, the market state, and each model's trailing realised IC lagged by $h$ days) is known at $t$. We test equal weighting, stacked ridge, stacked XGBoost, an MLP gate over models driven by the regime and skill, skill weighting, and a ridge-shrunk regime tilt of the equal-weight combination. The resulting layered system has four layers: \textbf{L1}, base forecasts; \textbf{L2}, equal-weight rank combination; \textbf{L3}, a regime exposure gate (a walk-forward ridge regression of the combination's realised IC on the market state; trade only when the predicted IC is positive); \textbf{L4}, a buffer rule (enter the top/bottom 20\%, keep while within the top/bottom 50\%).

\section{Results on the Development Period}
\label{sec:results}

\begin{table}[t]
\centering
\caption{Single stock (AAPL), corrected protocol: test period \aTestStart{} to \aTestEnd{} (\aNTest{} origins), 5-day horizon. Mean $\pm$ 95\% CI over five seeds. Sharpe ratio net of 5\,bp per side. DM: Diebold--Mariano $p$-value against the zero forecast (seed-averaged forecast, squared error).}
\label{tab:aapl}
\scriptsize
\setlength{\tabcolsep}{2.5pt}
\resizebox{\columnwidth}{!}{%
\begin{tabular}{lccccc}
\toprule
Model & MAE (\%) & $R^2_{\mathrm{OS}}$ (\%) & DA (\%) & Sharpe & DM $p$ \\
\midrule
Zero forecast & 3.12 & $-$1.9 & -- & -- & -- \\
Train mean (buy \& hold) & 3.08 & 0.0 & 57.2 & 0.99 & 0.466 \\
Ridge & 3.09 & $-$1.5 & 57.9 & 0.97 & 0.854 \\
XGBoost & 3.12\,$\pm$\,0.03 & $-$2.1\,$\pm$\,1.5 & 52.4\,$\pm$\,6.2 & 0.32\,$\pm$\,0.89 & 0.939 \\
LSTM & 3.78\,$\pm$\,0.45 & $-$41.2\,$\pm$\,23.9 & 48.4\,$\pm$\,3.0 & $-$0.14\,$\pm$\,0.57 & 0.016 \\
DLinear & 3.72\,$\pm$\,0.29 & $-$41.6\,$\pm$\,15.7 & 50.8\,$\pm$\,4.1 & $-$0.26\,$\pm$\,0.72 & 0.009 \\
Transformer & 3.41\,$\pm$\,0.30 & $-$16.1\,$\pm$\,14.2 & 47.8\,$\pm$\,4.4 & $-$0.35\,$\pm$\,0.52 & 0.065 \\
AFT (full) & 3.15\,$\pm$\,0.07 & $-$3.5\,$\pm$\,4.0 & 51.9\,$\pm$\,4.6 & 0.28\,$\pm$\,0.58 & 0.710 \\
AFT $-$ both attention biases & 3.15\,$\pm$\,0.07 & $-$3.6\,$\pm$\,4.2 & 51.7\,$\pm$\,4.7 & 0.29\,$\pm$\,0.59 & 0.697 \\
AFT $-$ regime encoder \& gate & 3.26\,$\pm$\,0.19 & $-$9.3\,$\pm$\,10.3 & 47.7\,$\pm$\,6.9 & $-$0.25\,$\pm$\,1.02 & 0.130 \\
AFT-R (renormalised gate) & 3.19\,$\pm$\,0.11 & $-$5.1\,$\pm$\,5.6 & 50.1\,$\pm$\,5.7 & 0.18\,$\pm$\,0.65 & 0.466 \\
\bottomrule

\end{tabular}%
}
\end{table}

\textbf{Single stock: no learned edge; inert components} (Table~\ref{tab:aapl}). The training-mean forecast (equivalent to buy-and-hold) has MAE \aMeanMae\% and a net Sharpe ratio of \aMeanSr; directional accuracy (DA) \emph{must} be compared with its \aMeanDa\%. LSTM, GRU and DLinear are significantly \emph{worse} than the zero forecast (DM $p=\aLstmDmZeroP$, \aGruDmZeroP, \aDlinDmZeroP). The AFT has the lowest error among the neural models (MAE \aAftMae\%) but no edge over the zero forecast (DM $p=\aAftDmZeroP$) and a Sharpe ratio of only \aAftSr. Removing both attention biases leaves the AFT unchanged (paired $t$-test over seeds: MAE $p=\aAblBothMaeP$, Sharpe $p=\aAblBothSrP$), and so does removing the regime encoder and gate ($p=\aAblRegMaeP$, \aAblRegSrP) or renormalising the gate ($p=\aAblAftrMaeP$, \aAblAftrSrP). The AFT's edge over the stock \texttt{nn.TransformerEncoder} model (MAE \aTrfMae\%) therefore comes from implementation details of its attention block, not from its financial components.

\begin{table*}[t]
\centering
\caption{Mean daily rank IC (Newey--West $t$-statistic) of base models and factors. Development period (2016--2022) for a hand-picked 100-stock universe and for the point-in-time S\&P~500, and the sealed holdout (2023--2025). Sharpe: quintile long-short, net of 5\,bp per side, S\&P~500, $h=21$, development period.}
\label{tab:base}
\footnotesize
\begin{tabular}{lcccccc}
\toprule
& \multicolumn{4}{c}{Development 2016--2022} & \multicolumn{2}{c}{Holdout 2023--2025} \\
\cmidrule(lr){2-5}\cmidrule(lr){6-7}
Model & 100 stocks, $h=5$ & S\&P~500, $h=5$ & S\&P~500, $h=21$ & Sharpe, $h=21$ & S\&P~500, $h=5$ & S\&P~500, $h=21$ \\
\midrule
Ridge & 0.007 (1.0) & 0.012 (2.0) & 0.030 (2.7) & 0.58 & 0.013 (1.4) & 0.018 (0.9) \\
XGBoost & 0.006 (0.8) & 0.015 (2.0) & 0.034 (2.5) & 0.55 & 0.021 (2.3) & $-$0.021 ($-$1.3) \\
MLP & 0.008 (1.1) & 0.010 (1.6) & 0.033 (2.0) & 0.34 & 0.012 (1.1) & 0.003 (0.2) \\
Transformer & -- & -- & 0.032 (1.6) & 0.32 & -- & $-$0.018 ($-$0.6) \\
AFT & -- & -- & 0.033 (1.8) & 0.36 & -- & $-$0.019 ($-$0.8) \\
AFT-R & -- & -- & 0.032 (1.7) & 0.35 & -- & $-$0.016 ($-$0.6) \\
5-day reversal & 0.013 (1.5) & 0.016 (2.0) & 0.021 (2.3) & 0.29 & 0.009 (1.0) & 0.005 (0.5) \\
21-day reversal & 0.009 (0.9) & 0.020 (2.3) & 0.037 (2.7) & 0.46 & 0.010 (1.0) & $-$0.009 ($-$0.5) \\
12-1 momentum & 0.009 (0.7) & $-$0.003 ($-$0.3) & $-$0.033 ($-$1.8) & $-$0.56 & 0.018 (1.3) & 0.018 (0.9) \\
Low volatility & $-$0.011 ($-$0.9) & $-$0.005 ($-$0.4) & $-$0.024 ($-$1.1) & $-$0.52 & $-$0.029 ($-$1.7) & $-$0.047 ($-$1.5) \\
MAX (lottery) & $-$0.011 ($-$1.1) & 0.001 (0.2) & $-$0.008 ($-$0.6) & $-$0.51 & $-$0.024 ($-$1.8) & $-$0.038 ($-$1.5) \\
\bottomrule

\end{tabular}
\end{table*}

\textbf{Breadth beats architecture} (Table~\ref{tab:base}). On 100 large caps no model is significant (e.g.\ XGBoost IC \tfXgbIc, $t=\tfXgbT$). On the point-in-time S\&P~500 the same models become detectable at $h=5$ (XGBoost \sfXgbIc, $t=\sfXgbT$; ridge \sfRidgeIc, $t=\sfRidgeT$), and more so at $h=21$ (XGBoost \stXgbIc, $t=\stXgbT$; ridge \stRidgeIc, $t=\stRidgeT$). Part of this is statistical power: each daily IC is estimated from a cross-section about four times wider, as the fundamental law suggests. The higher IC levels indicate that the broader universe also contains more predictable cross-sectional variation. In the 100-stock universe, neither rank versus $z$-score normalisation nor the original 95 features changed the picture significantly.

\textbf{The AFT equals a vanilla Transformer.} At $h=21$ the AFT's IC (\stAftIc) differs from the Transformer's (\stTrfIc) by \stAftVsTrfD{} ($p=\stAftVsTrfP$), and AFT-R (\stAftrIc) tracks the AFT almost exactly year by year. All three sequence models are less stable than the tabular models ($t\le\stAftT$; net Sharpe \stAftSr{} vs \stXgbSr{} for XGBoost; maximum drawdown \stAftMdd\% vs \stXgbMdd\%). Adding them to the combination lowered its net Sharpe ratio, so they were not used.

\begin{table}[t]
\centering
\caption{Learned meta-learners vs equal-weight rank combination (eight inputs: XGBoost, ridge, MLP and five factors), meta out-of-sample 2017--2022, S\&P~500. $p$: Newey--West test of the daily IC difference against equal weighting.}
\label{tab:meta}
\scriptsize
\setlength{\tabcolsep}{3pt}
\resizebox{\columnwidth}{!}{%
\begin{tabular}{lcccc}
\toprule
& \multicolumn{2}{c}{$h=5$} & \multicolumn{2}{c}{$h=21$} \\
\cmidrule(lr){2-3}\cmidrule(lr){4-5}
Meta-learner & IC & $p$ & IC & $p$ \\
\midrule
Equal-weight ranks (reference) & 0.0208 & -- & 0.0301 & -- \\
Regime-tilted EW (ridge) & 0.0033 & 0.080 & 0.0139 & 0.453 \\
Stacked ridge & $-$0.0018 & 0.090 & 0.0186 & 0.644 \\
Stacked XGBoost (+ skill, state) & 0.0033 & 0.117 & 0.0093 & 0.329 \\
Regime/skill-gated stacking (MLP gate) & 0.0029 & 0.071 & 0.0028 & 0.177 \\
Skill-weighted (63-day IC) & $-$0.0049 & 0.003 & $-$0.0114 & 0.004 \\
Skill-weighted (252-day IC) & 0.0084 & 0.002 & 0.0185 & 0.250 \\
\bottomrule

\end{tabular}%
}
\end{table}

\textbf{Learned meta-learning does not beat equal weighting} (Table~\ref{tab:meta}). None of six learned combiners outperforms equal weighting at either horizon. Weighting models by trailing realised skill is significantly \emph{worse}: skill is anti-persistent at the 63-day window. This is the forecast-combination puzzle~\cite{rapach2010} in a cross-sectional setting, and it is the meta-level counterpart of the inert AFT gate. With this little signal, estimated conditional weights add more noise than information.

\textbf{Regime dependence is real, but in a different place.} On the development period the IC of 5-day reversal rises steeply with market stress (it is concentrated in the top terciles of VIX, market volatility and cross-sectional dispersion; HAC slope $p\le0.031$), as liquidity-provision theory predicts~\cite{nagel2012}. Using this as an exposure gate helped reversal alone, but not the combination (Table~\ref{tab:system}).

\textbf{Execution matters.} The buffer rule cuts the combination's turnover at $h=5$ from \bufQTurn{} to \bufBTurn{} per rebalance and raises its net Sharpe ratio from \bufQSr{} to \bufBSr{} (at 10\,bp: \bufQSrt{} to \bufBSrt). In the 100-stock universe, by contrast, smoothing the signal with an EMA lost more predictive power than it saved in costs.

\section{Pre-registered Holdout}
\label{sec:holdout}
\begin{table*}[t]
\centering
\caption{Layered meta system on the point-in-time S\&P~500: development (2018--2022; the exposure gate needs one year of history) and sealed holdout (2023--2025). Inputs: XGBoost, ridge, MLP, 5- and 21-day reversal. Sharpe ratios of the long-short portfolio net of costs per side; turnover per rebalancing (2 = the whole book is replaced). The first row is the input with the best development Sharpe ratio.}
\label{tab:system}
\footnotesize
\setlength{\tabcolsep}{4pt}
\begin{tabular}{lcccccccc}
\toprule
& \multicolumn{4}{c}{Development 2018--2022} & \multicolumn{4}{c}{Holdout 2023--2025} \\
\cmidrule(lr){2-5}\cmidrule(lr){6-9}
System & IC & Sharpe 5\,bp & Sharpe 10\,bp & Turnover & IC & Sharpe 5\,bp & Sharpe 10\,bp & Turnover \\
\midrule
\multicolumn{9}{l}{\textit{Horizon $h=5$ days}} \\
Best single input (dev: 21-d rev.) & 0.017 & 0.30 & 0.11 & 1.56 & 0.010 & $-$0.08 & $-$0.38 & 1.51 \\
L2: EW combination & 0.020 & 0.33 & 0.03 & 2.41 & 0.017 & 0.10 & $-$0.40 & 2.43 \\
L2 + regime gate & 0.020 & 0.27 & 0.02 & 1.89 & 0.017 & 0.38 & $-$0.03 & 1.73 \\
\textbf{System A}: L2 + buffer & 0.020 & 0.39 & 0.20 & 1.22 & 0.017 & 0.34 & 0.05 & 1.17 \\
L2 + gate + buffer & 0.020 & 0.31 & 0.12 & 1.14 & 0.017 & 0.62 & 0.32 & 1.05 \\
System B + buffer & 0.019 & 0.35 & 0.16 & 1.19 & 0.012 & 0.14 & $-$0.15 & 1.15 \\
\midrule
\multicolumn{9}{l}{\textit{Horizon $h=21$ days}} \\
Best single input (dev: Ridge) & 0.036 & 0.76 & 0.63 & 2.55 & 0.018 & 0.46 & 0.35 & 2.35 \\
L2: EW combination & 0.040 & 0.59 & 0.48 & 2.86 & 0.000 & $-$0.02 & $-$0.15 & 2.82 \\
L2 + regime gate & 0.040 & 0.65 & 0.55 & 2.72 & 0.000 & $-$0.48 & $-$0.59 & 1.95 \\
\textbf{System A}: L2 + buffer & 0.040 & 0.59 & 0.51 & 1.70 & 0.000 & 0.08 & $-$0.02 & 1.76 \\
L2 + gate + buffer & 0.040 & 0.66 & 0.58 & 1.71 & 0.000 & $-$0.33 & $-$0.42 & 1.40 \\
System B + buffer & 0.036 & 0.53 & 0.46 & 1.59 & 0.002 & 0.13 & 0.04 & 1.72 \\
\bottomrule

\end{tabular}
\end{table*}

\begin{figure}[t]
\centering
\includegraphics[width=\columnwidth]{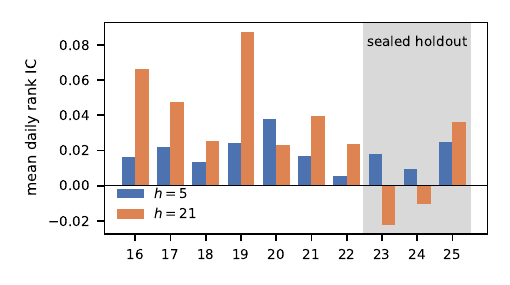}
\caption{Mean daily rank IC of the System A signal (equal-weight ranks of five inputs) by year, S\&P~500. Bars to the right of 2022 are the sealed holdout.}
\label{fig:icyear}
\end{figure}

Before the holdout was opened, we fixed the following: System A (equal-weight ranks of XGBoost, ridge, MLP and 5- and 21-day reversal, buffer rule) at $h=21$ as the primary system, because $h=21$ was clearly better on the development period (net Sharpe \systAdevSrf{} vs \sysfAdevSrf{} at $h=5$); two primary hypotheses (IC $>0$; net Sharpe at 5\,bp $>0$); and, as secondary results, the 5-day system, the gated variants, all base models and the replication of the reversal--regime relation.

\textbf{The primary hypothesis was not supported.} At $h=21$ the holdout IC is \hTIc{} ($t=\hTT$, one-sided $p=\hTP$; by year \hTIcYa, \hTIcYb, \hTIcYc) and the net Sharpe ratio is only \systAholdSrf{} at 5\,bp (positive, as the second hypothesis requires, but negligible) and \systAholdSrt{} at 10\,bp (Table~\ref{tab:system}, Fig.~\ref{fig:icyear}). The 21-day machine-learning signals reversed in the 2023--24 market, which was led by a few large technology stocks: XGBoost \htXgbIc, AFT \htAftIc, Transformer \htTrfIc{} (Table~\ref{tab:base}), while 12-1 momentum, negative over the development period, turned positive (\htMomIc). Ridge regression stayed positive (\htRidgeIc; net Sharpe \htRidgeSr).

\textbf{The 5-day system survived.} At $h=5$, System A has a holdout IC of \hFIc{} ($t=\hFT$, one-sided $p=\hFP$), positive in each year (\hFIcYa, \hFIcYb, \hFIcYc). Its net Sharpe ratio is \sysfAholdSrf{} at 5\,bp (gross \sysfAholdSrz, maximum drawdown \sysfAholdMdd\%), close to its development value (\sysfAdevSrf). With the exposure gate it reaches \sysfFullholdSrf{} at 5\,bp and \sysfFullholdSrt{} at 10\,bp, with the long-short portfolio active on \sysfFullholdExp\% of the rebalancing dates. XGBoost alone replicates its development IC (\hfXgbIc, $t=\hfXgbT$). These are secondary results: we report them as evidence, not as confirmed findings.

\textbf{The reversal--regime relation did not replicate.} On the holdout, none of the state variables significantly predicts the 5-day reversal IC (smallest slope $p=\sfMinP$; VIX $p=\sfVixP$). So the gate's holdout gain at $h=5$ cannot be attributed to the mechanism identified on the development period.

\section{Discussion and Limitations}
\textbf{What the evidence says.} The AFT's inductive bias does not fail because regimes are irrelevant. It fails because (i) its gate is attenuated by construction and, once repaired, still has nothing to exploit at the feature-group level; and (ii) with daily price and volume data the predictable component of returns is small, so a richer architecture mainly fits noise. The robust gains come from \emph{breadth}, from \emph{simple combination} and from \emph{cost-aware execution}. The meta-level evidence mirrors the architecture: learned gates and learned combination weights add estimation noise faster than they add information.

\textbf{What the holdout teaches.} The development period favoured $h=21$ by a wide margin, yet only $h=5$ held up. A single three-year holdout cannot say whether that is luck or structure; it does show how large the gap between development and out-of-sample performance can be, even under a careful protocol.

\textbf{Limitations.} The universe consists of today's S\&P~500 members (survivorship bias), with Yahoo Finance adjusted prices; costs are proportional, with no market impact or borrowing fees; the holdout spans only three years (about 36 non-overlapping 21-day periods per phase); and the 5-day result is a secondary analysis that requires confirmation on a new period (e.g.\ 2026 onward) or a different market. Hyper-parameters of the zoo were fixed a priori rather than tuned. Tuning could change individual models but would add further researcher degrees of freedom.

\section{Conclusion}
Under leakage-controlled, pre-registered evaluation, regime-gated attention adds nothing over a vanilla Transformer for stock-return prediction: its gate is attenuated by construction and its components are inert. Value instead comes from cross-sectional breadth, equal-weight combination of diverse forecasts, and cost-aware execution. The design that looked best on the development period failed on the sealed holdout, while its 5-day counterpart survived. We hope the released protocol helps future architecture papers in this area report results that survive out of sample.

\appendices
\section{Corrections Relative to Version~1}
\label{app:v1}
Version~1 of this preprint (``Adaptive Financial Transformer with Regime-Gated Attention for Stock Return Prediction'') reported results that this version replaces. An audit of the v1 notebooks found the following errors; all of them are corrected here, and all results in this version were regenerated by a single reproducible pipeline.
\begin{itemize}
\item The 95\% confidence intervals reported for the linear, random-forest, XGBoost, LSTM and GRU baselines were not produced by any experiment: each baseline had been trained once.
\item The row labelled GRU contained the metrics of a vanilla Transformer; the baselines' trading metrics came from an earlier backtest that compounded overlapping 5-day returns.
\item The component-ablation numbers do not appear in any experiment output. Under the corrected protocol, removing the AFT's components changes nothing (Table~\ref{tab:aapl}).
\item The main AAPL dataset covered 2015--2025 (2{,}721 rows), not 2018--2024 (1{,}762 rows); the reported sequence realignment was not used in the main benchmark; no transaction costs were applied although 0.05\% was stated; the final models were trained with mean-squared error, not with the described composite loss; the normalisation constants of the hyper-parameter objective were test-set values; the hyper-parameter search space, the best trial and the hardware were misreported; and the multi-stock study used undocumented hyper-parameters, different features and a single seed.
\item On the v1 test window the zero forecast had a lower error than both reported AFT configurations, and an always-long forecast had a higher directional accuracy; v1 did not report these benchmarks.
\end{itemize}
A detailed, itemised audit with evidence is in the repository (\texttt{AUDIT.md}).

\section{Reproducibility}
\label{app:repro}
All experiments ran on one NVIDIA GeForce RTX 4060 Laptop GPU with PyTorch 2.11.0 (CUDA 12.8), scikit-learn 1.9.0 and XGBoost 3.2.0, with deterministic CUDA kernels wherever PyTorch provides them (the vanilla Transformer's fused attention backward pass is not deterministic). The repository contains the feature pipeline with its look-ahead test, the walk-forward runner, the research log of all development iterations (\texttt{RESEARCH\_LOG.md}), the pre-registration (\texttt{PREREGISTRATION.md}), all summary results and the log of the single holdout evaluation. The out-of-sample predictions and model checkpoints are attached to release v2.0.0 of the repository (\url{https://github.com/Dishan18/FinancialTransformer/releases/tag/v2.0.0}), and every release of the code is archived on Zenodo (\url{https://doi.org/10.5281/zenodo.21109641}).

\end{document}